# Redefining Safety for Autonomous Vehicles

Philip Koopman✉[0000-0003-1658-2386], William Widen

Carnegie Mellon University, Pittsburgh PA, USA
University of Miami – School of Law, Miami FL, USA
`koopman@cmu.edu, wwiden@law.miami.edu`

**Abstract.** Existing definitions and associated conceptual frameworks for computer-based system safety should be revisited in light of real-world experiences from deploying autonomous vehicles. Current terminology used by industry safety standards emphasizes mitigation of risk from specifically identified hazards, and carries assumptions based on human-supervised vehicle operation. Operation without a human driver dramatically increases the scope of safety concerns, especially due to operation in an open world environment, a requirement to self-enforce operational limits, participation in an ad hoc sociotechnical system of systems, and a requirement to conform to both legal and ethical constraints. Existing standards and terminology only partially address these new challenges. We propose updated definitions for core system safety concepts that encompass these additional considerations as a starting point for evolving safety approaches to address these additional safety challenges. These results might additionally inform framing safety terminology for other autonomous system applications.

## 1    Introduction

The proliferation of critical systems that employ machine learning-based technology requires a fresh look at what we mean for a computer-based system to be "safe." It is time to ask whether our current definitions of safety are, themselves, fit for purpose. Existing definitional frameworks can be stretched further, but are showing significant signs of wear. Moreover, the mindset and processes informed by those legacy definitions are themselves eroding under the complexity of this new technology.

An update to the definition of safety and related terms can promote a more robust view of safety. We ground our proposed updates in incidents already seen in real-world driverless Autonomous Vehicle (AV) deployments as well as incidents involving human-supervised driving automation features, identifying broader underlying themes. We propose an updated set of definitions based on a need to evolve terminology used in current relevant industry consensus standards: ISO 26262, ISO 21448, and UL 4600.

As an initial example of the types of issues that are arising, there have been numerous reports of robotaxis in San Franciso interfering with emergency responder operations. This often does not involve overtly dangerous vehicle motion such as a crash, but rather might involve a vehicle immobilizing itself in a bid to mitigate the risk of a crash in a situation it cannot handle safely. However, such safety-motivated behavior can increase societal risk by impeding the progress of emergency response vehicles. This is not the type of loss event primarily contemplated by current safety terminology.

We use examples of recent vehicle automation incidents to identify shortcomings of current safety concepts and terminology. We then propose adjusted definitions for core safety terms. While we focus on terminology and vehicle automation in this paper, this is the start of a much larger discussion regarding the safety of autonomous systems for a variety of application domains. The nature of that new technology demands a major shift and expansion of safety perspective to deal with these challenges.

Section 2 of this paper summarizes current definitions for safety and related concepts, using international standards relevant to automated vehicle safety. In subsequent sections, we cover why those definitions struggle under the demands placed upon them by autonomous systems, and identify core gaps illustrated by real-world incidents. We then propose updated key definitions to address the gaps while not breaking existing safety engineering practices, staying as close to existing definitions as is practicable.

Out of scope for this paper are: organizational safety philosophy, how to best achieve a defined safety goal, and how to integrate engineering terms into law. We leave for other discussions important topics such as safety improvement approaches and safety management systems. Similarly, hazard identification and analysis techniques remain applicable but beyond our scope. We also acknowledge that there is extensive work on safety and risk in the legal domain which informs our work. However, we do not seek to propose a redefinition of legal terminology—a topic for another paper.

## 2    Existing safety definitions

At its heart, our work is an inquiry into changing the nature of what "safety" should mean when developing a computer-based safety-critical system – especially one that must operate within a sociotechnical framework that goes beyond the bounds of both traditional functional safety and human-supervised system safety.

This work builds on our initial work finding that acceptable autonomous vehicle safety requires more than a net risk "better than human driver" approach [14]. In that work, we identify complicating factors including: risk subsidy, risk transfer, negligence, responder role safety, the proper role of blame, standards conformance, regulatory requirements, ethical concerns, equity concerns, and per-behavior regulatory approaches to risk acceptability. That work encompasses various contributions from [1], [2], [5], [23], and [26], but does not consider the question of whether the language of risk and safety itself should be revisited to help resolve the issues raised.

We start with an examination of key definitions of safety from autonomous vehicle safety standards. An overarching concept of existing definitions in the automotive domain is that automotive feature safety is often considered to mean an ***Absence of Unreasonable Risk (AUR)***, typically as described in a chain of definitions in ISO 26262.

## 2.1   ISO 26262

ISO 26262:2018 is a functional safety standard that applies to road vehicles, whether autonomous or not [9]. That standard's Part 1 safety definitions are based on layers of risk, hazards, and harm. The core stack of definitions is summarized below, omitting cross-references and supporting notes. Defined terms are italicized:

- **Safety:** absence of *unreasonable risk*
- **Unreasonable risk:** *risk* judged to be unacceptable in a certain context according to valid societal moral concepts
- **Risk:** combination of the probability of occurrence of *harm* and the *severity* of that *harm*
- **Severity:** estimate of the extent of *harm* to one or more individuals that can occur in a potentially *hazardous event*
- **Hazardous event:** combination of a *hazard* and an *operational situation*
- **Hazard:** potential source of *harm* caused by *malfunctioning behavior* of the *item*
- **Harm:** physical injury or damage of persons
- **Malfunctioning behavior:** *failure* or unintended behavior of an *item* with respect to its design intent
- **Operational situation:** scenario that can occur during a vehicle's life
- **Safety case:** argument that *functional safety* is achieved for *items* or *elements*, and satisfied by evidence compiled from *work products* of activities during development.

We stop at this point without diving deeper into other defined terms. For our purposes, an "item" is an autonomous vehicle including its runtime support infrastructure, although it could be interpreted in other ways according to the standard. A failure is related to an abnormal condition, including both random and systematic faults, caused by component malfunctions, software defects, and any other sources.

As a functional safety standard, the emphasis is on avoiding harm to people due to malfunctioning behaviors, omitting mention of other types of potential losses such as property damage. The approach implicitly assumes that a system that perfectly implements its design intent (which includes risk mitigation measures) will be safe.

In practice, the automotive industry tends to identify hazards, analyze the risk of each hazard, and perform mitigation to ensure that no individual hazard results in an unreasonable risk. If that criterion is met, the system is deemed safe.

While it might happen in practice, there are no explicit requirements to examine relationships between risks presented by multiple unrelated hazards, to quantify net risk presented by the system as a whole, or to consider risks due to sources other than malfunctioning behaviors that violate design intent. In practice, a vehicle is presumed to be safe when it is released into series production. Any safety defect discovered after release is, in essence, due to a defective development or manufacturing process that might lead to a regulatory recall or other deployment of a manufacturer remedy.

The Automotive Safety Integrity Level (ASIL) approach considers severity, exposure, and controllability to determine the mitigation required for each identified hazard. Addressing  controllability for uncrewed vehicles is potentially problematic.

## 2.2    ISO 21448

ISO 21448:2022 covers Safety of the Intended Function (SOTIF) [10], expanding the definition of safety from ISO 26262 while serving as a complementary standard. Its scope encompasses the driving behaviors of autonomous vehicles.

ISO 21448 does not change the definition of risk, nor of safety, instead adopting the ISO 26262 definitions. It does, however, add a new term covering SOTIF. Again, definitions are listed with supporting material omitted and defined terms italicized:

- **Safety of the intended functionality (SOTIF):** absence of *unreasonable risk* due to *hazards* resulting from *functional insufficiencies* of the *intended function* or its implementation
- **Hazard:** potential source of harm caused by the hazardous behavior at the vehicle level
- **Functional insufficiency:** *insufficiency of specification* or *performance insufficiency*
- **Performance insufficiency:** limitation of the technical capability contributing to a hazardous behavior or inability to prevent or detect and mitigate reasonably foreseeable indirect *misuse* when activated by one or more *triggering conditions*
- **Operational design domain (ODD):** specific conditions under which the system is designed to function

Again we stop short of listing the entire linked set of defined terms.

SOTIF overall deals with two types of functional insufficiencies generally beyond the scope of ISO 26262: (1) incomplete requirements, and (2) technical limitations of the system. A functional insufficiency might produce hazardous behavior in a specific circumstance, known as a triggering condition.

The incomplete requirements aspect is intended to deal with the fact that public roads are complex operational environments. The SOTIF methodology is based on reducing the "unknown hazardous" scenario categories until AUR has been achieved ([10] figures 8 and 10). The possibility of requirements gaps is acknowledged via reference to a potential insufficiency of specification. While this standard contemplates that an iterative experimental approach will be required for hazard analysis and scenario identification, it presumes that complete-enough specifications can be developed to achieve AUR before deployment. Its definition of hazard is modified from that in ISO 26262 to emphasize dangerous vehicle-level behavior.

Technical limitations of any system with external sensors will inevitably require building an internal model of the external world based on limited, incomplete, and noisy information. Not every radar pulse will produce a return. Sensors do not have an infinite range. Some objects are occluded from the point of view of any particular sensor. And various types of noise will introduce uncertainty into any model of the outside world. This standard recognizes that an automated driving system must be designed for AUR despite those potential performance insufficiencies.

Again, the emphasis is on the vehicle's behavior, and a general assumption is that loss events will primarily result from crashes caused by sensor limitations, vehicle behavioral deficiencies, or specification insufficiencies.

The inclusion of the concept of a defined Operational Design Domain (ODD) raises an important point in terms limiting the scope of safety considerations. In classical terminology, safety tends to be limited to some defined ODD scope. For example, the definition of a safety case in DefStan 00-56 is limited to cover "a given application in a given operating environment" [4]. However, the consideration of how operation is limited to that defined scope is often deferred to the judgment of human operators, such as a pilot who is supposed to avoid flying in extreme environmental conditions. With an autonomous system, a limitation of safety to defined conditions is inadequate, because some actor (by default the autonomous system) must also ensure safety by avoiding operating outside the ODD. This might include refusing to start a mission outside the ODD. But it should also require some acceptably safe response to being forcibly ejected from the ODD on short notice due to an unforeseen condition or event. Thus, enforcement of the ODD must be within the scope of a relevant definition of safety.

### 2.3   ANSI/UL 4600

UL 4600 is a system-level safety standard specific to autonomous vehicles [27]. Its terminology is intended to be compatible with ISO 26262 and ISO 21448, while addressing a broader system-level scope. Definitions were methodically reconciled with safety standards from other domains and are summarized below:

- **Safe:** having an *acceptable* post-mitigation *risk* at the *item* level as defined by the *safety case*
- **Acceptable:** sufficient to achieve the overall *item risk* as determined in the *safety case*
- **Risk:** combination of the probability of occurrence of a *loss* event and the severity of that *loss* event
- **Loss:** a substantive adverse outcome, from damage to property or the environment, to animal injury or death, to human injury or death
- **Safety case:** structured argument, supported by a body of evidence, that provides a compelling, comprehensible, and valid case that a system is safe for a given application in a given environment

As with previously discussed standards, risk is related to the probability and severity of a specific loss event. However, the notion of acceptability of risk has to do with an overall item risk rather than individual risks, including for example a highly recommended approach of total item risk summing (UL 4600 prompt element 6.1.1.3.a). Additionally, the definition of a loss is expanded past harm to humans to include other types of adverse outcomes, such as property damage.

The scope of a safety case is broadened, but its interaction with the definition of the term "safe" gives latitude to the authors of the safety case. It is the responsibility of the creator of the safety case to define what "safe" might mean, and to ensure that the safety case provides a suitable argument showing that the goal of safety has been achieved. The source for the definition of 'safety case' is Def Stan 00-56 [4], which limits the scope of the safety case to predetermined applications in a given operational environment. However, UL 4600 provides extensive lists of prompt elements to encourage robust consideration of potentially exceptional aspects of the ODD.

### 2.4    Other safety definitions

Other automotive safety standards in development that we are aware of plan to adopt ISO 26262 and potentially ISO 21448 terminology in whole or in part.

Another definition of safety widely cited in the automotive industry is Positive Risk Balance (PRB). This is one safety consideration proposed in a BMVI report [1]. That report proposes other restrictions on ethical safety, such as avoiding using personal characteristics to choose victims in a no-win crash scenario. But PRB is typically the criterion singled out for broader automotive industry safety discussions.

Using PRB as a sole criterion is problematic in no small part because of the difficulties in establishing a comparable baseline for human-driven vs. autonomous vehicles [14]. Nonetheless, it is the primary criterion promoted by both Waymo [29] and Cruise [3] when messaging their safety. PRB is often presumed to yield AUR in such discussions, although examples of AV safety problems discussed in the next section suggest this will not necessarily be the case.

An important regulatory user of the concept of AUR is the US National Highway Traffic Safety Administration (NHTSA). NHTSA pursues enforcement action when "the Agency finds either non-compliance or a defect posing an unreasonable risk to safety" [20]. Their approach tends to have two elements. The first is compliance with the Federal Motor Vehicle Safety Standards (FMVSS) [21], a set of specific tests for specific safety features primarily applicable to conventional vehicle safety rather than automated vehicle functions. FMVSS is necessary but not sufficient for safety.

The second NHTSA criterion is that they consider a safety defect to be a specific behavior, design defect, or other issue that they can associate with a pattern of loss events. NHTSA can launch investigations and require safety recalls. Typical NHTSA recalls and investigations involve a failure to conform to FMVSS or, less often, patterns of mishaps and incidents that are more difficult to link to a specific technical defect. NHTSA decisions have not historically considered net PRB at the vehicle level.

## 3    Examples of AV safety problems

In this section we consider some of the many real-world incidents and mishaps suffered by robotaxis to provide grounding for identifying gaps in current safety terminology. Examples are drawn from robotaxis and other vehicle automation incidents. Whether a safety supervisor human driver was present for an incident is irrelevant—a safety issue that happened with a human safety driver might also happen without one.

**Pedestrian dragging.** A pedestrian was hit by another vehicle and thrown into the path of a robotaxi [12]. The robotaxi braked aggressively but struck the pedestrian. Arguably the robotaxi could have driven more defensively to perhaps avoid that initial strike. Regardless, after having stopped, the robotaxi lost track of the pedestrian and decided to pull to the side of the road, dragging the pedestrian under the vehicle, ending up with the pedestrian pinned almost entirely under the rear of the vehicle. The robotaxi company involved attempted to portray this as an unforeseeable freak event. Nonetheless, moving a vehicle without first ascertaining the location of an injured pedestrian who has just been struck by that same vehicle is highly problematic.

**Crash with a firetruck.** A robotaxi entered an intersection as permitted by a green light but then collided with a fire truck, resulting in a passenger injury [16]. The fire truck had emergency annunciators active (siren, lights, horn) and was proceeding through a red light in cross traffic while responding to an emergency call. The robotaxi failed to yield to that emergency vehicle as required by road rules.

**Hitting a bus.** A robotaxi became confused when following a long bus with a mid-body articulation pivot [24]. The robotaxi tracked the front half of the bus and ignored the back half. It then crashed into the back half of the bus because the tracking system had decided to ignore the detected back half in favor of the front half.

**Interfering with emergency responders.** City of San Francisco emergency responders reported at least 55 incidents of robotaxi interference with their operations [7], later increased to 74 incidents reported by the fire department [8]. While no incidents were definitively shown to result in harm to a person, they presented risk by delaying emergency responders and requiring attention from emergency response personnel that would be better spent tending to the actual emergency.

**Encroachment on closed roads.** There have been numerous incidents of encroachment on closed roads and emergency scenes which presented potential hazards to vehicles, vehicle occupants, and other road users. Examples include: dragging downed power lines and emergency scene yellow barrier tape down the street [24], and driving through a construction zone only to get mired in wet concrete [12].

**Mass strandings.** Numerous mass strandings of vehicles have occurred, under a variety of circumstances. One that got particular attention was attributed to a loss of communications due to cellular phone system overload by a concert event in a proximate geographic region – even though that event did not involve the street on which the stranding took place [18]. This raises questions as to what would happen in a communication disruption or traffic control device power outage caused by a natural disaster such as an earthquake or other common cause infrastructure failure.

**Child debarking a school bus.** A vehicle with driving automation activated struck and injured a child debarking from a school bus [17].

**Failure to stop at stop signs (rolling stops).** A NHTSA recall for safety defects was implemented for a driving automation system that was programmed to roll through stops at speeds up to 5.6 mph, in violation of traffic laws [28].

**Emergency responder injuries and fatalities.** An investigation and initial recall were conducted for a pattern of collisions with emergency response vehicles that, over time, involved at least 14 crashes, 15 injuries, and 1 fatality [22]. An eventual recall was not for a specific reproducible behavioral defect, but rather a pattern of losses with a common theme related to lack of ODD enforcement combined with inadequate enforcement of human driver attention on the road [11].

**Collisions with stopped and crossing vehicles.** Related to the emergency responder crash investigation are numerous reports of collisions with stopped vehicles that have been reported for a particular driving automation system. That includes multiple fatalities in scenarios involving under-running a crossing heavy truck (e.g. [25]). While the manufacturer says that the design intent is for the driver to manually avoid such driving situations and such collisions, crashes keep accumulating nonetheless.

**Elevated collisions in a vulnerable community.** A concerning 11 out of 74 San Francisco Fire Department reported robotaxi incidents, including a collision with a fire truck, occurred in the vicinity of the geographically small Tenderloin district [8]. This district is renowned for being disadvantaged and having a historically at-risk community. For likely related reasons, it is one of the most active emergency response locations in the US. Nonetheless, robotaxi companies have seen fit to continue testing in that area, presumably because of its location in downtown San Francisco.

**Attracting passengers away from mass transit.** Even if robotaxis were as safe as human-driven vehicles, human-driven vehicles are far more dangerous than mass transit [19]. A widespread adoption of robotaxis could degrade safety by shifting mass transit passenger-miles to robotaxi passenger-miles [31]. Loss of mass transit patrons could additionally erode the funding and viability of safer modes of transportation, increasing net fatalities totaled across all modes of transportation.

We acknowledge that human drivers can and do make all of the types of mistakes recounted above. But we are interested in understanding the true scope of safety, which should apply to both human drivers and vehicles with automated driving features.

Many of these situations did not involve actual harm to a person. But all were considered safety issues by at least some relevant stakeholders due to the potential for direct or indirect harm. In many of the above cases it is easy to blame some actor other than the driving automation capability for responsibility. But casting blame to dodge changing the status quo is unlikely to prevent future mishaps.

## 4    What is missing from safety definitions

Based on these observed failures and a general understanding of autonomous vehicle safety requirements, we identify four general characteristics of autonomous vehicles that profoundly affect safety engineering: operating in an open world environment, self-enforcing operational limits, deployment in an ad hoc sociotechnical system, and addressing external constraints such as legal limitations. Dealing with all four of these areas has historically been allocated to the human vehicle driver. However, the whole point of having an autonomous vehicle is to no longer need that human driver, imposing these additional requirements on technical systems instead. (As an interim measure, remote operations teams might assist with some of these issues. But scalable deployment requires minimizing the need for such remote human operator intervention.)

### 4.1    Open world environment

Autonomous vehicles are engineered systems that must operate within a framework of uncertainty and incomplete training for all of the objects and events they will encounter while operating on public roads at scale.

The SOTIF approach of ISO 21448 is largely intended to address this issue. However, that standard's approach – and the practical approach of many developers – is to assume that enough possible scenarios, objects, and events will have been identified and mitigated before deployment to result in net AUR. This might not be possible in a practical system if the distribution frequency of encountering hazards is heavy-tailed, involving a large number of individual hazards that each have a very low frequency of arrival [13]. While there are possible technical approaches to ensure acceptable safety despite a heavy-tailed hazard situation, the probability of substantive risk from such hazards remaining after deployment cannot be ignored. ISO 21448 accommodates an iterative improvement approach to deal with newly emergent hazards, but still presumes that AUR will be achieved at initial release.

UL 4600 has more comprehensive mechanisms that recognize a substantive degree of uncertainty might be present at initial deployment, meaning AUR might be expected, but that expectation might itself have a substantial degree of uncertainty. UL 4600 requires the use of Safety Performance Indicators and field engineering feedback to manage a continuous improvement process to identify and mitigate risks due to a changing, open world environment as well as encounters with unforeseen heavy-tail events.

A comprehensive definition of safety should contemplate two issues that will be a reality for autonomous vehicles for the foreseeable future: (1) proactive management of inevitable requirements gaps in deployed systems, and (2) support for continual updates over the vehicle's lifecycle to mitigate emergent hazards and risks due to environmental and other changes.

## 4.2    Self-enforcement of operational limitations

A common approach to reaching a situation outside the operational limits of an AV is some sort of safety shutdown or abnormal mission termination. For an AV this might mean pulling to a safe stopping location, or even stopping in the middle of a travel lane in favorable conditions. While executing a reasonable safety stop can be complex, recognizing that the AV has exceeded its operational limitations is even more challenging.

Machine learning-based technology has a fundamental challenge with recognizing that it has encountered a meaningful data dimension that is relevant to safety but has not been captured as a data feature of some sort during its training. For example, if there are too few people dressed in yellow clothing in a training data set, a yellow-garbed construction worker might not be recognized as a construction worker directing traffic, or might not even be recognized as a person at all. Or emergency scene yellow tape might be recognized as an insubstantial bit of plastic that poses no collision threat instead of constituting a safety-relevant "keep-out" warning.

A core element of safety will need to be recognizing and responding to situations that exceed the intended nominal operational environment of the system. These might be unforeseen situations such as novel objects and events. But they might also be an unexpected arrival of a foreseen out-of-ODD situation, such as a sudden torrent of rain on a day with a sunny weather forecast. This couples with the open environment issue to make it challenging to self-enforce operational limits inherent to the system that were not contemplated as such by the design team. Current definitions tend to limit safety to being considered within an understood, specified environment. However, autonomous systems must also ensure safety when operating in an under-specified environment, as well as be able to react in some reasonable way to unexpectedly finding themselves outside the environment they were designed to operate within.

### 4.3    Ad hoc systems of systems

Autonomous systems must operate as a component of societal systems, which are often under-specified and, for the most part, beyond the ability of the autonomous system design team to control. From the point of view of current safety definitions, the design team cannot control which scenarios it will have to deal with other than by, in some cases, establishing operational limitations that the system must then self-enforce.

A common theme of some stakeholder safety concerns is that an autonomous vehicle is acting in a narrowly safe way by not crashing into things, but is causing negative externalities for other road users. Examples include an in-lane stop when the AV is unsure what to do next, missing cues that an apparently open driving lane (intended for construction vehicle use) should not be used by a robotaxi due to the context of "road closed" signs in adjacent lanes, a practical necessity to break some normal road usage practices in exceptional circumstances to provide room for a passing emergency vehicle, elevated risk of crashes due to so-called "phantom" braking, etc.

Some safety concerns have a more subtle context sensitivity. For example, a human driver might reroute due to seeing a huge structure fire a few blocks ahead to avoid becoming ensnarled with the likely on-scene traffic chaos. A robotaxi that proceeds without recognizing the situation will potentially impede emergency response activity.

While one might try to analyze all the hazards present at the system-of-systems level, this is typically beyond the scope and resources of an AV safety engineering effort. We believe it is more practical to instead express mitigation for negative externalities as constraints on permissible behavior. An example rule might be to pull out of travel lanes when a fire truck with active annunciators is in the vicinity regardless of on-board software's estimate of the risk of collision. Another example might be to avoid using a road where a preceding robotaxi has gotten stuck to avoid clustering stuck robotaxis.

### 4.4    Legal and ethical constraints

There are a number of constraints on permissible system behavior that not only are difficult to express as hazards, but might reduce the theoretical net benefits (or possibly even net safety) of a system. A risk-centric approach will struggle with such constraints.

As an example, consider a hypothetical situation in which a driving automation system reduced total fatalities, but increased the rate of fatalities imposed on emergency responders at roadside crash scenes. Or consider a more extreme hypothetical situation in which total road fatalities were reduced by half—but pedestrian fatalities doubled in number, becoming a much bigger fraction of that reduced net fatality rate. Both outcomes will be problematic for some stakeholders, even if net harm is reduced.

A technical system should not be considered acceptably safe by societal stakeholders if it does not satisfy constraints on both individual vehicle behaviors and patterns of behaviors that would, for example, also apply to human drivers in such situations. Consider a vehicle that rolls through stop signs when it determines the intersection is clear. Perhaps, hypothetically, a study might show that this reduces rear-end collisions and therefore improves net safety. Such a system would still likely be seen as presenting unreasonable risk due to automating the violation of traffic laws. It would also present a negligent driving liability exposure if an AV were to hit an undetected pedestrian after ignoring the requirement to come to a full and complete stop at a stop sign.

There are also a number of ethical, equity, and legal concerns that are not obviously tied to malfunctioning vehicle motion control, such as concerns of over-weighting public road testing of immature technology in vulnerable communities [30]. Some such constraints can be converted to functional requirements, but others might better be treated as constraints on design optimization choices [15].

## 5    A proposal for more robust definitions

### 5.1    What needs to be addressed?

We propose an updated set of core safety definitions to address the issues identified based on examples of vehicle automation incidents, in light of concepts already present in the standards and other sources mentioned in the preceding sections.

Reviewing the example incidents and analysis of gap areas in sections 3 and 4 above, we believe that the following aspects of safety need to be addressed more directly at the definitional level. We use keywords and phrases from definitions surveyed in section 2 as tags for each concept.

- **Safe/Safety:** The system must not only mitigate hazards, but also meet externally imposed constraints. Constraints might involve ethical- and equity-based prohibitions on engineering optimizations that might otherwise improve specific aspects of safety [15], and also prohibit unacceptable patterns of risks. We prefer "has **acceptable safety**" to "safe."
- **ODD/Given environment:** The environment cannot be assumed to be fully characterized, nor unchanging over time. Rather, acceptable safety must be assured regardless of the real-world environment, even if the system must self-enforce a risk mitigation response due to exceeding its operational limitations.
- **Given application:** The system needs to enforce potential misuse, such as an operator attempting to engage driving automation on roads with cross-traffic when that might violate a limited access highway-only system design intent restriction.
- **Risk:** The time-worn formulation of risk as a combination of probability and severity might serve for single-dimensional optimization of net harm, especially in situations in which monetary compensation is a morally acceptable plan for mitigation. But it is an overly narrow viewpoint for societal and other constraints such as patterns of harm or violations of constraints not readily reduced to a classical risk value, especially when a purely utilitarian approach might be deemed undesirable.
- **Hazardous event:** There are some risks and constraints that are challenging to evaluate at the hazardous event level because they involve safety tradeoffs at the system-of-systems level and potentially unknown requirements (think of future case law that has not yet been established). It is unreasonable to expect vehicle designers to fully assess, let alone mitigate every single such risk up front before systems are deployed. Patterns of risk such as risk transfer onto vulnerable populations are rather far removed from individual loss events.

- **Severity:** Harm done by any particular loss event is important, but the severity of any individual event might not capture the importance of that event if it is part of a larger pattern of unethical or inequitable outcomes – even if that event would otherwise be considered low severity in terms of utilitarian personal harm.
- **Malfunctioning:** Incidents can occur not just because a vehicle has displayed dangerous motion, but also because in an attempt to improve tactical safety it inflicts

---

- **Acceptable:** meets all *safety constraints* as shown by a *safety case*
  *Note: The phrase "acceptably safe" might be used in some contexts for clarity. While "Safety" is used as a modifier, use of the word "safe" alone should be avoided. Safety constraints encompass to whom the safety must be acceptable.*
- **Safety case:** structured argument, supported by a body of evidence, that provides a compelling, comprehensible, and sound argument that *safety engineering* efforts have ensured a system meets a comprehensive set of *safety constraints*
  *Note: This emphasizes meeting constraints rather than net risk. A limit to defined operational environments is intentionally excluded, but ODD enforcement might be allocated to human operators in the safety case when appropriate.*
- **Safety engineering:** a methodical process of ensuring a system meets all its *safety constraints* throughout its lifecycle, including at least hazard analysis, risk assessment, risk mitigation, validation, and field engineering feedback
  *Note: Requires safety engineering beyond brute force test validation. Hazard analysis is broadened to address all safety constraints. Explicitly requires addressing safety over the system's lifecycle.*
- **Safety constraint:** a limitation imposed on *risk* or other aspects of the system by stakeholder requirements
  *Note: This implicitly requires the identification of stakeholders who might be affected by losses, and makes it more straightforward to view safety as a multi-dimensional constrained optimization problem rather than a mostly one-dimensional pure risk optimization problem [15]. Safety constraints might include: AUR, PRB, limits on individual risks, limits on net risk, exposure limits for specified types of risk patterns, and issues that are difficult to trace to pure risk.*
- **Risk:** combination of the probability of occurrence of a *loss*, or pattern of *losses*, and the importance to stakeholders of the associated consequences
  *Note: Consequence (severity) might be an overriding concern regardless of probability. Net importance can be non-linearly related to individual losses if forming a pattern. Correlated loss events, inequitable loss patterns, and loss patterns involving a failure to mitigate emergent loss trends are in-scope.*
- **Loss:** an adverse outcome, including damage to the system itself, negative societal externalities, damage to property, damage to the environment, injury or death to animals, and injury or death to people
  *Note: This is broader in scope than some other typical definitions of loss or harm. Some types of loss might be assigned very low severity in some application domains. Allocation of blame does not affect whether a loss occurred.*

*Figure 1. Proposed definitions.*

potential damage at the system-of-systems level. Immobilized robotaxis blocking emergency vehicles after a safety shutdown are a poster child of this issue.

- **Harm:** Harm must be considered to go beyond personal injury or fatalities (as is already the case in some definitions). Even indirectly caused harm to property might legitimately be seen as a safety issue for some domains.

### 5.2    Proposed safety-related definitions

We propose a set of new definitions for core safety terminology in figure 1 above. While specific proposals for each standard's particular use of terms are beyond the scope of this paper, we believe that if these terms were to be adopted by UL 4600 the changes imposed on the remainder of the document would guide a beneficial evolution to the scope of coverage of safety cases conforming to that standard. Other standards such as ISO 26262 might still appropriately use more restrictive terms, but should align terminology so as not to preclude conformance if these more expansive definitions are used by design teams instead.

At a high level, the approach suggested here considers the concept of "safety" not as an optimization process to reduce risk, but rather as the satisfaction of a set of safety constraints. One such constraint will typically be sufficient risk mitigation in keeping with more traditional safety engineering approaches. However, other constraints together with an increased scope for the concept of a loss event can address legal and ethical issues. The open world environment issue is addressed by explicitly including lifecycle considerations in the definition of safety engineering, and requiring safety engineering to be tied to a safety case. The self-enforcement of operational limitations issue is addressed via removing the phrase "for a given environment" and avoiding reference to an ODD in the definition of the safety case. The ad hoc system of systems issue is addressed via including the notion of stakeholders beyond the system designer in the definitions of risk and safety constraint.

## 6    Conclusions

While there is a continual stream of new safety considerations in the evolution of the area of safety engineering, we believe that the advent of autonomous systems represents a watershed moment. Moreover, the concerns motivating the proposed changes are not just theoretical, but have actually played out on US public roads. This is just the beginning. It is time for the safety community to revisit what safety really means.

One might consider addressing the concerns we raise via aggressive reinterpretation of existing terminology. However, any compliance-centric users of a standard are strongly incentivized to interpret definitions in the way most favorable to a low-cost compliance effort, undermining any potential educational efforts to promote reinterpretation to expand definitional scope. Additionally, a standard should say what it means, and not be dependent on non-normative, independent interpretational guidance.

While the standards mentioned in this paper were all written in good faith and have served honorably, it is time to update their definitions of safety to address the concerns raised by real-world experiences with the increasing complexities of automation technology. To be clear, we consider these proposed definitions to be the start of a discussion within the safety community rather than a finished conclusive outcome.

While we have used autonomous vehicles as a motivating example, similar issues will arise across a broad spectrum of safety-critical systems, and these definition proposals might inform autonomous system safety more broadly. What has changed is the lack of a closely attentive human operator to address issues such as enforcing operational limits, or to potentially serve as a moral crumple zone [6] to shield the system from blame for unmitigated equipment failures. An additional factor to consider is the relationship between the definitions for safety and cybersecurity terminology, especially when security failures can compromise safety.

Broadening the scope of safety will be an improvement for all systems, and becomes increasingly important as technology continues to insinuate itself into the fabric of everyday society. The more we automate beyond a practical ability for humans to exercise effective oversight, the more pressing these issues will become.

Thanks to Dr. Mallory Graydon and the anonymous reviewers for their comments.

# 7    Appendix

Space limitations restrict the number of cross-references to existing definitions within the main paper. This Appendix provides summaries of additional definitions from freely available documents to provide further context. A serious definitional effort within a standards framework should additionally consider pay-to-view standards materials as well as standards still in the development process.

Free on-line access to the definition sections standards mentioned in the main body of the paper can be found as follows:

- ISO 26262: https://www.iso.org/obp/ui/#iso:std:iso:26262:-1:ed-2:v1:en
- ISO 21448:  https://www.iso.org/obp/ui/#iso:std:iso:21448:ed-1:v1:en
- Free Digital View of UL 4600: https://www.shopulstandards.com/ProductDetail.aspx?productid=UL4600

## 7.1    SafAD

*Reference: "Safety First for Autonomous Driving" whitepaper*
*https://www.connectedautomateddriving.eu/wp-content/uploads/2019/09/Safety_First_for_Automated_Driving.pdf*

The SafAD white paper is the result of a collaboration among multiple companies to summarize autonomous vehicle design, verification, and validation methods. It proposes twelve foundational principles as well as functional elements for implementing an autonomous driving system

Key safety definitions from the section 7 glossary of this reference take ISO 26262 as a starting point:

- **Safe(ty):** absence of *unreasonable risk* due to hazards
- **Unreasonable risk:** risk judged to be un*acceptable* in a certain context according to valid societal moral concepts
- **Acceptable risk:** post-mitigation risk is acceptable to the developing company and also with respect to legal and social acceptance criteria
- **Accident:** an undesirable, unplanned event that leads to an unrecoverable loss of service due to unfavorable external conditions, typically involving material damage, financial loss, and harm to humans
- **Positive risk balance:** lower remaining risk of traffic participation due to automated vehicles, including fewer crashes on average compared to an average human driver

These definitions seem to give developers an outsized vote on determining acceptability, but do acknowledge that there are other legal and social acceptance criteria. The definition of accident includes loss of service, but does not address negative externalities explicitly, including perhaps external harm that does not impair vehicle operation.

Positive risk balance seems to be the safety objective (e.g., according to the abstract), but the link between that objective and the definition of safety could be clarified. An approach to tie positive risk balance to safety would likely use the key challenges identified in section 3.2 of that report as a starting point. Terms undefined in the SafAD paper would be expected to have definitions according to ISO 26262.

## 7.2    MIL STD 882E

*Reference: MIL-STD-882E w/CHANGE 1 27 September 2023*
*https://quicksearch.dla.mil/qsDocDetails.aspx?ident_number=36027*

MIL-STD-882E is a system safety standard from the US Department of Defense that applies to a variety of systems including autonomous ground vehicles. Due to its applicability to combat systems, it has some unique concepts compared to automotive and other industrial standards, such as differentiating harm to civilians vs. harm to combatants. However, core defined terms are quite relevant to this discussion.

- **Acceptable Risk:** *risk* that the appropriate acceptance authority is willing to accept without additional mitigation
- **Risk:** a combination of the *severity* of the *mishap* and the *probability* that the *mishap* will occur
- **Severity:** magnitude of potential consequences of a *mishap*, including death, injury, occupational illness, damage to or loss of equipment or property, damage to the environment, or monetary loss
- **Probability:** an expression of the likelihood of occurrence of a *mishap*
- **Mishap:** event or series of events resulting in unintentional death, injury, occupational illness, damage to or loss of equipment or property, or damage to the environment
- **Safety:** freedom from conditions that can cause death, injury, occupational illness, damage to or loss of equipment or property, or damage to the environment

While safety is defined, the high-level goal of safety engineering within MIL-STD-882E is to achieve acceptable risk. The definition of acceptable risk traces to a definition of a mishap, which recites the same criteria included in the definition of safety.

In this standard acceptability is determined by an identified risk acceptance authority, rather than by reference to more generic industry or societal norms. This is a reflection of unique tradeoffs that must be made for systems intended for use in combat situations, which might include accepting unmitigated risk due to an otherwise favorable net loss/benefit analysis, especially for an urgently needed combat capability.

## 7.3    DefStan 00-56

*Reference: DefSTan 00-56 Part 1, issue 7, 28 Feb 2017.*
*Free access is available via: https://www.dstan.mod.uk/StanMIS/Account/Login?ReturnUrl=%2fStanMIS*

DefSTan 00-56 is a UK Ministry of Defense standard for system safety, with terminology defined in Part 1, Appendix A. A key deliverable for this standard is a safety case.

- **Safety Case:** a structured argument, supported by a body of evidence that provides a compelling, comprehensible, and valid case that a system is *safe* for a given application in a given operating environment
- **Safe:** freedom from unacceptable or intolerable levels of *harm*
- **Risk:** combination of the likelihood of *harm* and the *severity* of that *harm*
- **Severity:** A measure of the degree of *harm*

- **Harm:** adverse impact on people, including fatality, physical or psychological injury, or short or long term damage to health
- **ALARP:** As Low As Reasonably Practicable (as clarified in other MoD guidance)

In this standard, the notion that levels of harm might be unacceptable or intolerable corresponds to the ALARP concept, in which risks must be mitigated until they are either tolerable or acceptable. In some ways, this is a more rigorous criterion than AUR, because it requires risk mitigation beyond AUR if doing so is practicable. However, it presents a serious potential safety gap for new technology for which there might not be any practicable way to mitigate risk to a level that is acceptable to relevant stakeholders. In short, ALARP might be seen as either (or both) too rigorous and too lax for novel or immature technologies compared to other acceptance criteria.

The definition of safety case includes a limitation to a "given operating environment," which generally corresponds to the notion of an ODD for autonomous vehicles. This definition should be updated to ensure safety for autonomous systems that must detect their own departures from the intended operating environment.

## 7.4    ISO 14971:2019

*Reference: ISO 14971:2019(en) Medical devices — Application of risk management to medical devices*
*https://www.iso.org/obp/ui/#iso:std:iso:14971:ed-3:v1:en*

ISO 14971 provides a framework for manufacturers to systematically manage the risks associated with the use of medical devices. Key definitions include:

- **Safety:** freedom from unacceptable *risk*
- **Risk:** combination of the probability of occurrence of *harm* and the *severity* of that *harm*
- **Severity:** measure of the possible consequences of a *hazard*
- **Hazard:** a potential source of *harm*
- **Harm:** injury or damage to the health of people, or damage to property or the environment

Rather than specifying a particular operational environment, the medical domain frames the scope of safety within an "intended use" or "intended purpose". Additionally, the terms "risk management" and "risk control" are used rather than the risk mitigation phrasing commonly used in other standards. Nonetheless, the concepts map fairly well to other standards we discuss.

An issue analogous to enforcement of an ODD arises with determining whether a medical device is serving an "intended use" or "intended purpose" rather than a "reasonably foreseeable misuse" (all of these are defined terms in this standard). An autonomous medical device would need to enforce boundaries of potential misuse (intentional or otherwise) for situations beyond the reasonable control of a human medical practitioner to monitor or enforce.

### 7.5    The Functional Safety Terminology Landscape

Reference: *The Functional Safety Terminology Landscape, 30 Jan. 2023.*
*https://ieeexplore.ieee.org/servlet/opac?punumber=10042203*

This is a survey of existing terminology across numerous standards, including many that are not discussed above. It compares and contrasts various definitions and recommends a net outcome set of definitions, including those listed below.

- **Acceptable:** "unacceptable" and "acceptable" are used in discussions, but not directly defined. However, unreasonable and tolerable risks are defined:
  - *Unreasonable risk:* "risk judged to be unacceptable in a certain context according to valid societal moral concepts"
  - *Tolerable risk*: "risk judged to be acceptable in a certain context according to valid societal moral concepts"
- **Safety:** "freedom from losses considered to be unacceptable in a defined context"
- **Safety case:** not defined.
- **Safety engineering:** not defined, but a relevant supporting definition is provided:
  - **Level of rigor:** "a specification of the depth and breadth of engineering activities necessary to provide a sufficient level of confidence that a safety-critical or safety-related function will perform as required"
- **Safety constraint:** not defined.
- **Risk:** "the effect of uncertainty on objectives resulting from the combination of the likelihood of occurrence of unwanted behavior and the effect of the unwanted behavior"
- **Loss:** not directly defined, but discussions use the terms loss and harm. A relevant supporting definition is:
  - **Hazard:** "potential source of harm that is a threat to safety"